\documentclass[10pt,twocolumn,letterpaper]{article}

\usepackage{cvpr}
\usepackage{times}
\usepackage{epsfig}
\usepackage{graphicx}
\usepackage{amsmath}
\usepackage{amssymb}
\usepackage[english]{babel}
\usepackage[utf8]{inputenc}
\usepackage{multirow}
\usepackage[linesnumbered,ruled,vlined]{algorithm2e}
\usepackage{array}
\usepackage{booktabs}

\makeatletter
\newcommand{\thickhline}{%
    \noalign {\ifnum 0=`}\fi \hrule height 2pt
    \futurelet \reserved@a \@xhline
}
\newcolumntype{"}{@{\hskip\tabcolsep\vrule width 2pt\hskip\tabcolsep}}
\makeatother

\usepackage[pagebackref=true,breaklinks=true,colorlinks,bookmarks=false]{hyperref}
\cvprfinalcopy 

\def\cvprPaperID{3} 

\newcommand*{\affaddr}[1]{#1} 
\newcommand*{\affmark}[1][*]{\textsuperscript{#1}}
\newcommand*{\email}[1]{\texttt{#1}}

\begin{document}

\title{Binarized Convolutional Neural Networks with Separable Filters \\
for Efficient Hardware Acceleration}

\author{%
Jeng-Hau Lin\affmark[1], Tianwei Xing\affmark[2], Ritchie Zhao\affmark[3], Zhiru Zhang\affmark[3], Mani Srivastava\affmark[2],\\ Zhuowen Tu\affmark[4,1] and Rajesh K. Gupta\affmark[1]\\
\affaddr{\affmark[1]Department of Computer Science and Engineering, UC San Diego}\\
\affaddr{\affmark[2]Department of Electrical Engineering, UC Los Angeles}\\
\affaddr{\affmark[3]Department of Electrical and Computer Engineering, Cornell University}\\
\affaddr{\affmark[4]Department of Cognitive Science, UC San Diego}\\
\email{\affmark[1]\{jel252,ztu,rgupta\}@ucsd.edu, \affmark[2]\{twxing,mbs\}@ucla.edu,}\\
\email{\affmark[3]\{rz252,zhiruz\}@cornell.edu}\\
}

\maketitle

\begin{abstract}
State-of-the-art convolutional neural networks are enormously costly in both compute and memory, demanding massively parallel GPUs for execution.
Such networks strain the computational capabilities and energy available to embedded and mobile processing platforms, restricting their use in many important applications.
In this paper, we push the boundaries of hardware-effective CNN design by proposing BCNN with Separable Filters (BCNNw/SF), which applies Singular Value Decomposition (SVD) on BCNN kernels to further reduce computational and storage complexity.
To enable its implementation, we provide a closed form of the gradient over SVD to calculate the exact gradient with respect to every binarized weight in backward propagation.
We verify BCNNw/SF on the MNIST, CIFAR-10, and SVHN datasets, and implement an accelerator for CIFAR-10 on FPGA hardware.
Our BCNNw/SF accelerator realizes memory savings of $17$\% and execution time reduction of $31.3\%$ compared to BCNN with only minor accuracy sacrifices.
\end{abstract}


\section{Introduction}
\label{sec:intro}


Albeit the community of neural networks has been prospering for decades, state-of-the-art CNNs still demand\\ 

\date{
	{\footnotesize \textbf{Acknowledgement}: This work was performed within the CRAFT project (DARPA Award HR0011-16-C-0037), and supported by NSF IIS-1618477, and a research gift from Xilinx, Inc. Any findings in this material are those of the author(s) and do not reflect the views of any of the above funding agencies. The U.S. and U.K. Governments are authorized to reproduce and distribute reprints for Government purposes notwithstanding any copyright notation hereon.}
}

\noindent  significant computing resources (i.e., high-performance GPUs), and are eminently unsuited for resource and power-limited embedded hardware or Internet-of-Things (IoT) platforms~\cite{rastegari2016xnornet}.
Reasons for high resource needs include the complexity of connections among layers, the sheer number of fixed-point multiplication and accumulation (MAC) operations, and the storage requirements for weights and biases.
Even if network training is done off-line, only a few high-end IoT devices can realistically carry out the forward propagation of even a simple CNN for image classification.

Binarized convolutional neural networks (BCNNs)~\cite{hubara2016bnn, courbariaux2015binaryconnect, soudry2014ebp, minje2016bitwise, rastegari2016xnornet} have been proposed as a more hardware-friendly model with extremely degenerated precision of weights and activations.
BCNN replaces floating or fixed-point multiplies with XNOR operations (which can be implemented extremely efficiently on ASIC or FPGA devices) and achieved near state-of-the art accuracy on a number of real-world image datasets at time of publication.
Unfortunately, this hardware efficiency is offset by the fact that a BCNN model is typically tens or hundreds times the size of a CNN model of equal accuracy.
To make BCNNs practical, an effective way to reduce the model size is required.

\begin{figure}
	\label{fig:ori_bina_SVD}
	\centering
	\includegraphics[width=0.7\columnwidth]{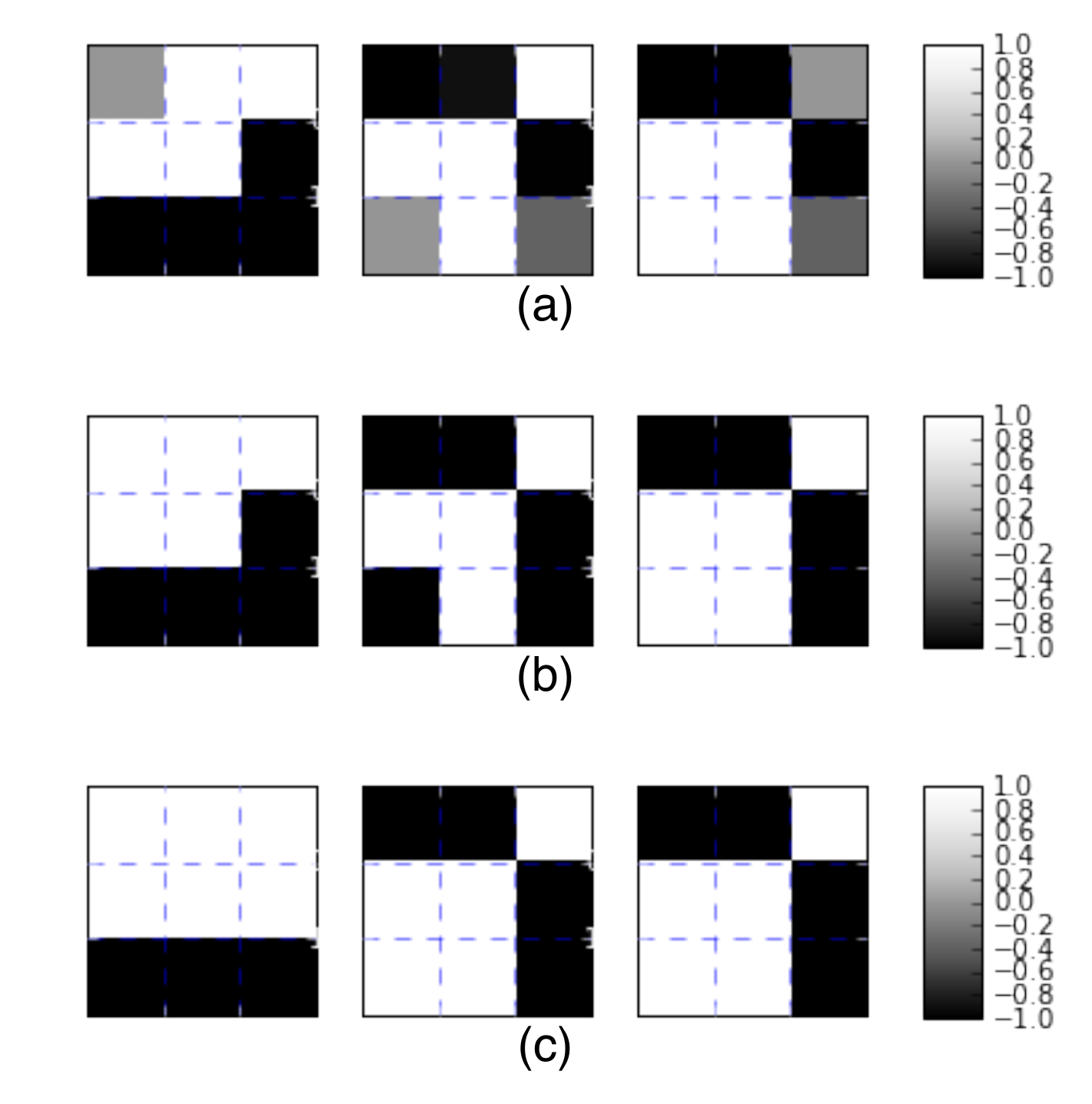}
	\caption{Comparison of filters: (a) original floating point filter; (b) same  filter binarized; (c) approximated separable binary filter.}
\end{figure}
In this paper, we introduce Separable Filters (SF) on binarized filters, as shown in Fig.~\ref{fig:ori_bina_SVD}(c), to further reduce the hardware complexity in two aspects:
\begin{itemize}
    \item SF reduces the number of possible unique $d$-by-$d$ filters from $2^{d^2}$ to just $2^{2d-1}$, enabling the use of a small look-up table during forward propagation. This directly results in the $\frac{(d-1)^2}{d^2}$ reduction of memory footprint.
    \item SF replaces each $d$-by-$d$ 2D convolution with two $d$-length 1D convolutions, which reduces the number of MAC operations by $d/2$. This translates to either speedup or the same throughput with fewer resources.
    \end{itemize}

In addition, we propose two methods to train BCNNw/SF: 
\\\textbf{Method 1 - Extended Straight-through Estimator (eSTE)}: take the rank-1 approximation for SFs as a process adding noise into the model and rely on batch normalization to regularize the noise. During backward propagation, we extend the straight-through estimator (STE) to propagate gradient across the decomposition. 
\\\textbf{Method 2 - Gradient over SVD}: go through the analytic closed form of the gradient over SVD to push the chain rule in backward propagation to the binarized filters, which is the filter before SVD.


The rest of the paper is organized as follows: Sec.~\ref{sec:related} provides a brief survey of previous works, Sec.~\ref{sec:BCNNSF} presents the design of BCNNw/SF and some implementation details, Sec.~\ref{sec:BPSF} presents two methods for the training of BCNNw/SF., Sec.~\ref{sec:experi} shows experimental results, Sec.~\ref{sec:FPGA} describes the implementation of BCNNw/SF on an FPGA platform, and Sec.~\ref{sec:concl} concludes the paper.

\section{Related Works}
\label{sec:related}

We leverage the lightweight method for training a BCNN as proposed by Hubara et al.~\cite{hubara2016bnn, courbariaux2015binaryconnect}, which achieved state-of-the-art results on datasets such as CIFAR-10 and SVHN.
Two important ideas contributed to the effectiveness of their BCNN:
\\\textbf{Batch normalization with scaling and shifting~\cite{ioffe2015batchnorm}}: A BN layer regularizes the training process by shifting the mean to zero, making binarization more discriminative. It also introduces two extra degrees of freedom in every neuron to further compensate for additive noises.
\\\textbf{Larger Model}: As with the well-known XOR problem~\cite{rumelhart1985errorprop}, using a larger network increases the power of the model by increasing the number of dimensions for projection and making the decision boundary more complex.

Rastegari et al. proposed XNOR-Net~\cite{rastegari2016xnornet}, an alternative BCNN formulation which relies on a multiplicative scaling layer instead batch normalization to regularize the additive noise introduced by binarization.
The scaling factors are calculated to minimize the 1-norm error between real-valued and binary filters.
While Hubara's BCNN did not perform well with a larger dataset such as ImageNet~\cite{deng2009imagenet}, obtaining a top-1 error rate of 72.1\%, XNOR-Net improves this error rate to 55.8\%.

Rigamonti et al.~\cite{rigamonti2013separable_filter} proposed a rank-1 approximate method to replace the 2-D convolution in a CNN with two successive 1-D convolutions.
Every filter was approximated by the outer product of a column vector and a row vector which were obtained through Singular Value Decomposition (SVD).
The authors proposed two schemes of the learning of separable filters: (1) retain only the largest singular value and corresponding vectors to reconstruct a filter; (2) linearly combine the outer products to lower the error rate.
However, the first scheme sacrificed too much performance because the the other singular values can be comparable with the largest one in terms of magnitude.
The second scheme was designed to compensate for loss of performance, but more singular values used to recover a filter means a lesser benefit from the approximation.
Although learning with separable filters was computationally expensive, the low rank approximation is an important idea to alleviate hardware complexity.

Inspired by Rigamonti's work, more research projects has been conducted to explore a more economic model, \ie networks with smaller memory requirements for the kernels.
Jaderberg~\etal~\cite{jaderberg2014low_rank} proposed a filter compression method that analyzed the redundancy in a pre-trained model, decomposed the filters into single-channel separable filters, and then linearly combined separable filters to recover original filters.
The decomposition was optimized to minimize the L2 reconstruction error of original filters.
Alvarez~\etal~\cite{Alvarez2016decomposeme} presented DecomposeMe that further reduced the redundancy by sharing the separated filters in the same layer.
To alleviate the computational congestion of GoogLeNet~\cite{szegedy2015googlenet}, Szegedy~\etal~\cite{szegedy2015inception_v2, szegedy2016inception_v4} proposed a multi-channel asymmetric convolutional structure, which has the same architecture as the second scheme in the work of Jaderberg~\etal~\cite{jaderberg2014low_rank} but in different purposes:
Szegedy used the asymmetric convolutional structure to avoid the expensive 2D convolutions and train the filter directly, while Jaderberg decomposed pre-trained filters to exploit both input and output redundancies.
However, both Jaderberg's and Alvarez's methods required a pre-trained model, and both Jaderberg's and Szegedy's multi-channel asymmetric convolution brought additional channels requiring a larger memory footprint.

Our proposed method differs from the three methods above because we maintain the network structure during training phase, train rank-1 separable filters directly, and then decompose the rank-1 filters into pairs of vector filters for hardware implementation.
Last but not least, to the best of our knowledge no existing work provides an analytic closed form of the gradient of filter-decomposition process for backward propagation.

\section{Binarized CNN with Separable Filters}
\label{sec:BCNNSF}

Here we describe the theory of BCNN with Separable filter in detail.
Our main idea is to apply SVD on binarized filters to further reduce the memory requirement and computation complexity for hardware implementation.
We present the details of forward propagation in this section and two methods of backward propagation in the next section.

\subsection{The Subject of Decomposition}
\label{sec:design_choice}
\begin{figure}
	\centering
	\includegraphics[width=0.7\columnwidth]{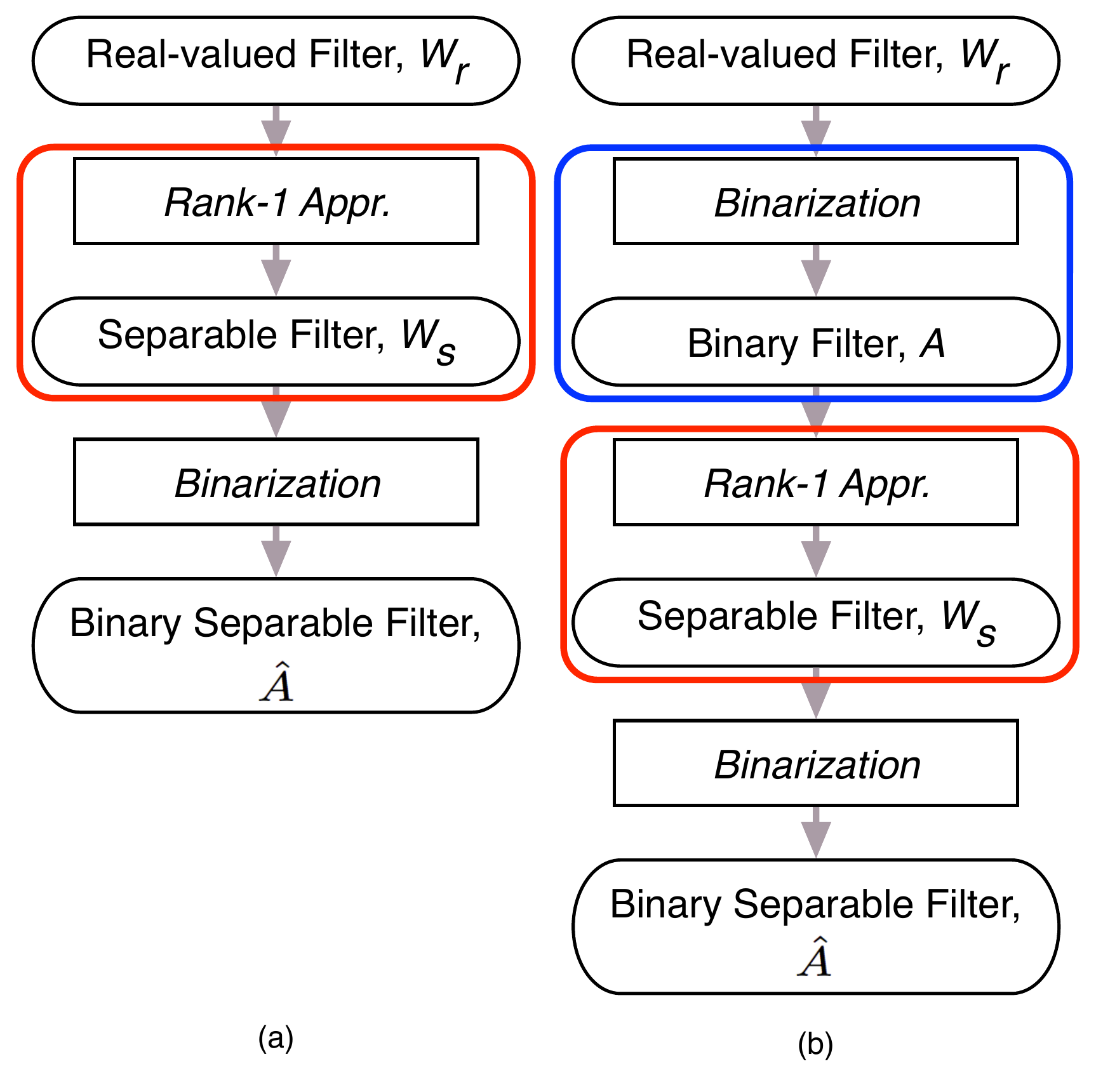}
	\caption{Comparison of the two SVD flows; (a) Flow 1: binarize the result of SVD on the floating-point filter; (b) Flow 2: directly decompose the binarized filters.}
    \label{fig:two_choices}
\end{figure}
For BCNN, there are two approaches to binary filter decomposition.
Fig.~\ref{fig:two_choices} depicts the two choices.
If we adopt flow 1 and apply the rank-1 approximation (the red box) directly on the real-valued filters, we cannot avoid real-time decomposition during training because the input filter has an infinite number of possible combination of pixel strengths.
Therefore, we introduce an extra binarization (the blue box) on the real-valued filters and apply the rank-1 approximation on the binarized filters.
Then, the number of possible input filters of rank-1 approximation are limited to $2^{d^2}$, where $d$ is the width or height of a filter.
With flow 2, we can build a look-up table beforehand and avoid real-time SVD during training.

Naturally, the rank-1 approximation and the extra binarization will limit the size of the basis to recover the original filters and equivalently introduce more noise into the model, as shown in Fig.~\ref{fig:ori_bina_SVD} from (b) to (c).
Instead of introducing an additional linear-combination layer to improve the accuracy, we leave the task to the two aforementioned reasons that make BNN work.




\subsection{Binarized Separable Filters}
Here we provide the detailed steps from binarized filters to binarized separable filters.
The result of SVD on a matrix $A$ includes three matrices as shown in Eq.~\ref{eq:SVD}.
\begin{equation}
\label{eq:SVD}
A = UDV^T
\end{equation}
Similar to real value rank-1 approximation for separable filter, the binarized separable filters are obtained with an extra binarization process on the dominate singular vectors as shown in Eq.~\ref{eq:rank_1_approx}.
\begin{equation}
\label{eq:rank_1_approx}
\hat A = b(U[:,1])b(V[:,1]^T),
\end{equation}
where $U[:,1]$ and $V[:,1]$ stand for the left and right singular vector corresponding to the largest singular value, respectively, and the function $b(.)$ denotes the binarization and can be implemented in either a deterministic function or a stochastic process~\cite{hubara2016bnn}.
Please note the largest singular value is dropped because all singular values are always positive and have no effect on binarization.


\begin{figure}
	\centering
	\includegraphics[width=0.6\columnwidth]{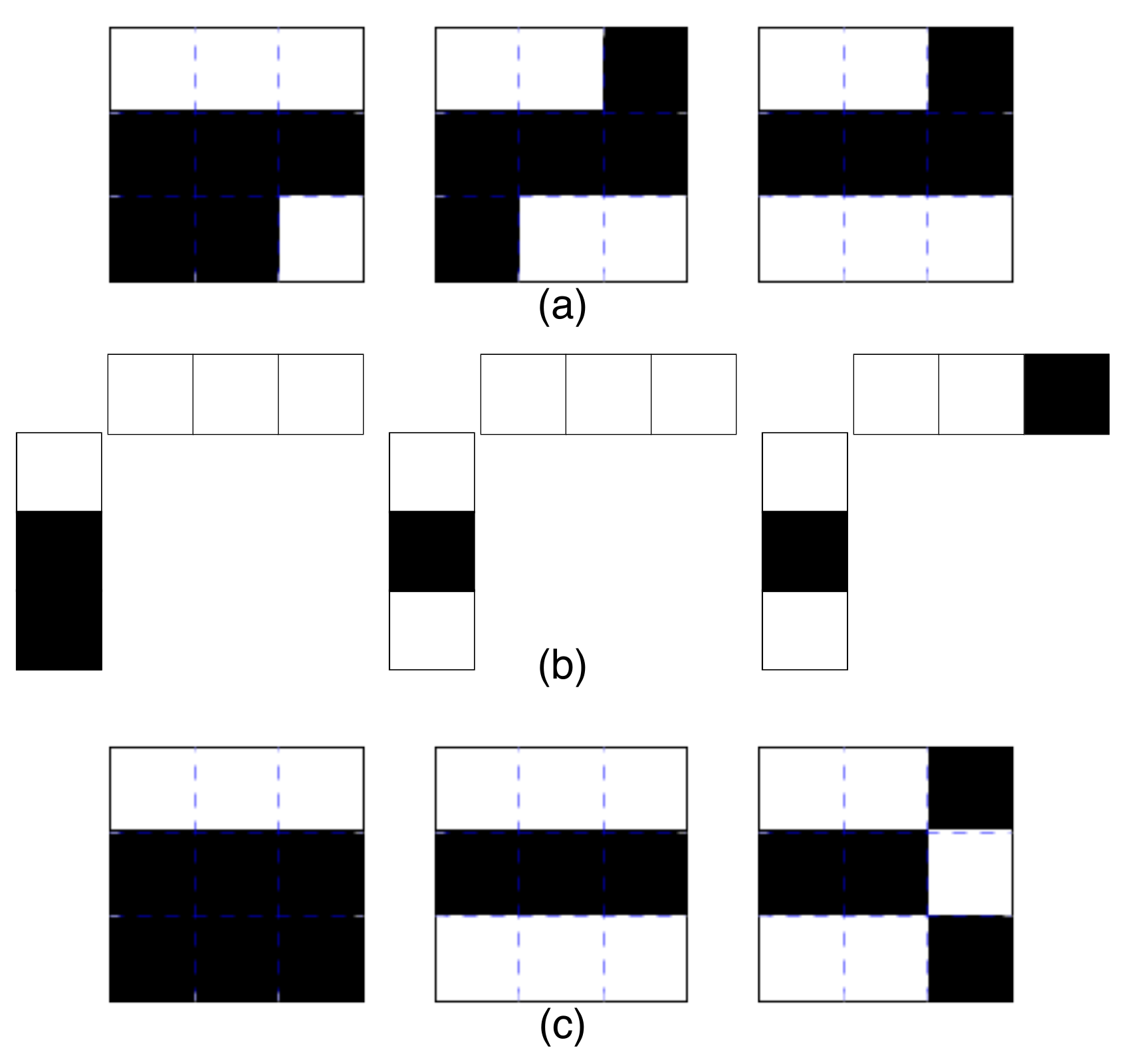}
	\caption{(a) a kernel before approximation; (b) pairs of vectors $(u,v^T)$ in SVD; (c) a kernel after rank-1 approximation in which every filter is an outer product of $u$, and $v$. The white and black colors stand for $+1$ and $-1$, respectively.}
	\label{fig:SVDkernel}
\end{figure}
 Fig.~\ref{fig:SVDkernel}(a) and (c) illustrates a kernel with three filters before and after binarized rank-1 approximation.
As with~\cite{hubara2016bnn} we keep a copy of the real-valued filters during each training iteration and accumulate the weight gradients on them since SGD and other convex optimization methods presume a continuous hypothesis space.
This also allows us to train the kernels as if the model is real-valued without the need for penalty rules~\cite{rigamonti2013separable_filter} during the backward propagation.  
For the test phase, all filters are binarized and rank-1 approximated to be binarized separable.

In our FPGA implementation, we use the pairs of vectors in Fig.~\ref{fig:SVDkernel}(b) to replace 2-D filters and perform separable convolution, which involves a row-wise 1D convolution followed by a column-wise convolution in back-to-back fashion before accumulating across different channels. More details on the FPGA implementation are presented in Sec.~\ref{sec:FPGA}.

\subsection{Details of the Implementation}
As mentioned in sec.~\ref{sec:design_choice}, the benefit of flow 2 is to leverage a finite-sized look-up table (LUT) to replace the costly SVD computation during the forward propagation of training phase.
Although the training takes place on a highly-optimized parallel computing machine, the LUT access is still a potential bottleneck if searching for an entry in the mapping is not efficient enough.

We build two tables to avoid real-time SVD.
The first table is composed by all binarized separable filters.
The number of entries in the first table can be calculated with Eq.~\ref{eq:num_rank1_filters}.
\begin{equation}
K = 2^{2d-1},
\label{eq:num_rank1_filters}
\end{equation}
where $d$ is the width or height of a filter.

The second table is the mapping relationship between all possible binary filters to their corresponding binarized separable filters.
We design an estimation function to make the tables content-addressable.
The key to index the first table  can be obtained with Eq.~\ref{eq:hash_fn}.
\begin{equation}
\label{eq:hash_fn}
key = \Lambda \cdot A,
\end{equation}
where $\Lambda$ is a vector or a matrix in the same size of $A$, and all elements in $\Lambda$ are the weightings to convert a matrix $A$ into a number.
The simplest choice of $\Lambda$ is the binary-to-integer conversion method.
We take the first element in $A$ as the least significant bit (LSB), so the $\Lambda$ is designed as Eq.~\ref{eq:Lambda}
\begin{equation}
\label{eq:Lambda}
\Lambda = \begin{bmatrix}
2^0 & 2^1 & 2^2 & \dots & 2^N
\end{bmatrix},
\end{equation}
where $N$ is the amount of elements of $A$, and $N=d^2$.
With this simple hash function and the efficient broadcasting technique in Theano~\cite{theano2016arXivshort}, we are able to efficiently obtain the keys for all filters in a convolutional layer.


\section{Backward Propagation of Separable Filters}
\label{sec:BPSF}
Besides the extra degrees of freedom introduced to BCNNw/SF's forward propagation, there are two more important techniques making binarized separable filters work.
In this section, we present two methods utilizing the two techniques for the training of BCNNw/SF.

\subsection{Method 1: Extended STE}
\label{sec:method_1}
As shown in Fig.~\ref{fig:two_choices}(b), during the forward propagation, all filters must be degraded thrice.
Since binarization can be considered as noise addition into the model and be regularized with batch normalization, the rank-1 approximation, which is just another process adding extra noise, can be regularized as well.
In details, we extend the straight-through estimator across the three degradation processes in Fig.~\ref{fig:two_choices} to update the real-valued filters with the rank-1 approximated filters.
Eq.~\ref{eq:extended_direct_est} shows the backward propagation of the gradient of rank-1 approximated filter,$g_{bs}$, to the gradient of real-valued filter, $g_r$.
\begin{equation}
g_r = g_{\hat A} \texttt{1}_{\left | r \right | \le \texttt{1}}
\label{eq:extended_direct_est}
\end{equation}
This simple method totally relies on batch normalization to regularize the noise introduced by two binarization and one rank-1 approximation.

\subsection{Method 2: Gradient over SVD}
\label{sec:method_2}
Whereas binarization is not a continuous function, Hubara~\etal~\cite{hubara2016bnn} resorted to the STE to update the real-valued weights with the gradient of loss w.r.t binarized weights.
Howbeit, owing to the continuity of singular value decomposition, we are allowed to calculate the gradient w.r.t. the resultant of the first binarization, $W_b$.
More specifically, the rank-1 approximation is differentiable because all of the three resultant matrices,~\ie $U$,$D$, and $V$, of SVD in Eq.~\ref{eq:SVD} are differentiable w.r.t. every element of the original input matrix, $A$. 
From the approximation we adopt for separable filters as shown in Eq.~\ref{eq:rank_1_approx}, one can easily obtain the derivative of $\hat A$ w.r.t. the elements of the original matrix before the approximation as Eq.~\ref{eq:diff_approx_A}, if the STE for binarization is applied.
\begin{equation}
\label{eq:diff_approx_A}
\frac{\partial \hat A}{\partial a_{ij}} = \frac{\partial U[:,1]}{\partial a_{ij}}b(V[:,1]^T) + 
b(U[:1]) \frac{\partial V[:,1]^T}{\partial a_{ij}}
\end{equation}

Papadopoulo~\etal~\cite{papadopoulo2000jsvd} provided the mathematical closed form of the gradient of the three resultant matrices, as shown in Eq.~\ref{eq:partial_u}, and \ref{eq:partial_v}.
\begin{eqnarray}
\label{eq:partial_u}
\frac{\partial U}{\partial a_{ij}} = U \Omega_U^{ij}\\
\label{eq:partial_v}
\frac{\partial V}{\partial a_{ij}}= -V \Omega_V^{ij},
\end{eqnarray}
where $\Omega_U^ij$ and $\Omega_V^ij$ are anti-symmetric matrices with zeros on their diagonals, and all off-diagonal elements can be obtained by Eq.~\ref{eq:omega_U} and ~\ref{eq:omega_V}.
\begin{eqnarray}
\label{eq:omega_U}
\Omega_{U_{kl}}^{ij} = \frac{d_l u_{ik} v_{jl} + d_k u_{il} v_{jk}}{d_l^2-d_k^2} \\
\label{eq:omega_V}
\Omega_{V_{kl}}^{ij} = \frac{d_k u_{ik} v_{jl}+d_l u_{il} v_{jk}}{d_k^2-d_l^2}
\end{eqnarray}

Eq.~\ref{eq:partial_hatA} shows the general form of the differential equation. 
\begin{equation}
\label{eq:partial_hatA}
\frac{\partial \hat A}{\partial a_{ij}} = 
\begin{bmatrix}
 \frac{\partial \hat a_{11}}{\partial a_{ij}}& \frac{\partial \hat a_{12}}{\partial a_{ij}} & \dots & \frac{\partial \hat a_{1N}}{\partial a_{ij}}\\ 
 \frac{\partial \hat a_{21}}{\partial a_{ij}}& \frac{\partial \hat a_{22}}{\partial a_{ij}} &\dots & \frac{\partial \hat a_{2N}}{\partial a_{ij}}\\ 
 \vdots &  \vdots & \ddots  &  \vdots\\ 
\frac{\partial \hat a_{M1}}{\partial a_{ij}}& \frac{\partial \hat a_{M2}}{\partial a_{ij}} & \dots & \frac{\partial \hat A_{MN}}{\partial a_{ij}}
\end{bmatrix}
\end{equation}
\begin{equation}
\label{eq:partial_hata}
\frac{\partial \hat a_{kl}}{\partial a_{ij}} = 
b(U_{k1})\sum_{n=2}^{N}{V_{kn}\Omega_{V_{1n}}^{ij}}-b(V_{l1})\sum_{n=2}^{N}{U_{ln}\Omega_{U_{1n}}^{ij}}
\end{equation}
From Papadopoulo's equations~\ref{eq:partial_u} to~\ref{eq:omega_V}, we can derive every element in Eq.~\ref{eq:partial_hatA} as shown in Eq.~\ref{eq:partial_hata} and see there exist cross-terms between elements.
The gradient of a SVD resultant matrix w.r.t. one element in the original input matrix is also a matrix of the same dimension, $M$by$N$, \ie a single element's change in the input matrix can affect all other elements in the resultant of SVD.
The intuition behind is that the rank-1 approximation is a matrix-wise filter-level mapping relationship rather than an element-wise operation, and multiple elements contribute to the mapping result of a filter.

To recap Eq.~\ref{eq:partial_hatA} with the chain rule calculation of backward propagation, we follow the similar fashion how higher layer neurons collect errors from the lower layer.
Eq.~\ref{eq:gOsvd_collect} shows the inner product for collecting error from lower layer and propagate the error to every element in binarized filters $A$.
For method 2, we also build a table of the derivatives together with the binarized rank-1 approximation to avoid real-time calculation of Eq.~\ref{eq:partial_hatA}.
\begin{equation}
\label{eq:gOsvd_collect}
\frac{\partial loss}{\partial a_{ij}} \equiv \frac{d loss}{d \hat A} \cdot \frac{\partial \hat A}{\partial a_{ij}}
\end{equation}

\section{Experiments}
\label{sec:experi}
We conduct experiments on the Theano~\cite{theano2016arXivshort} based on the Courbariaux's framework~\cite{courbariaux2016binarynetgit}, using 2 GPUs: NVIDIA GeForce GTX Titan X and GTX 970 to finish the training/testing process.
In most of the experiments, we obtain near state-of-the-art results using BCNNw/SF.

In this section, we describe the network structures we use, and list the classification result on $3$ datasets. 
We compare our result with relevant works, and then make analysis on different perspectives, including binarized separable filter and learning ripples.

\subsection{Datasets and Models}

We evaluate our methods on three benchmark image classification datasets: MNIST, CIFAR-10 and SVHN.
MNIST is a dataset for $28$x$28$ gray-scale handwritten digits, which has a training set of $60$K examples, and a testing set of $10$K examples.
SVHN is a real-world image dataset for street view house numbers, cropped to $32$x$32$ color images, with $604$K digits for training, $26$K digits for testing.
Both of these datasets classify digits ranging from $0$ to $9$.
CIFAR-10 dataset consists of $60$K $32$x$32$ color images in $10$ mutually exclusive classes (airplane, automobile, bird, cat, deer, dog, frog, horse, ship and truck), with $6,000$ images per class.
There are $50$K training images and $10$K test images.

The convolutional neural networks we use has almost the same architecture as Hubara~\etal~\cite{hubara2016bnn}'s except for some small modification.
This architecture is inspired from the VGG~\cite{simonyan2015vgg} network.
It contains $3$ fully-connected layers and $6$ convolutional layers, in which the kernels for convolutional layers is $3$ x $3$.
For detailed network structure parameters, see Table~\ref{tab:archi}.

\begin{table}[t]
\centering
\begin{tabular}{|c"c|c|c|}
\hline
Name       & MNIST(CNN)  & CIFAR-10     & SVHN       \\ \thickhline
Input      & 1x28     & 3x32x32     & 3x32x32     \\ \hline
Conv-1     & 64x3x3    & 128x3x3   & 64x3x3    \\
Conv-2     & 64x3x3   & 128x3x3 & 64x3x3   \\ \hline
Pooling    & \multicolumn{3}{c|}{2 x 2 Max Pooling}  \\ \hline
Conv-3     & 128x3x3  & 256x3x3 & 128x3x3  \\
Conv-4     & 128x3x3 & 256x3x3 & 128x3x3 \\ \hline
Pooling    & \multicolumn{3}{c|}{2 x 2 Max Pooling}  \\ \hline
Conv-5     & 256x3x3 & 512x3x3 & 256x3x3 \\
Conv-6     & 256x3x3 & 512x3x3 & 256x3x3 \\ \hline
Pooling    & \multicolumn{3}{c|}{2 x 2 Max Pooling}  \\ \hline
FC-1       & 1024   & 1024   & 1024   \\ \hline
FC-2       & 1024   & 1024   & 1024   \\ \hline
FC-3       & 10     & 10     & 10     \\ \hline
\end{tabular}
\vspace{0.2cm}
\caption{Network architecture for different datasets. 
The dimension of a convolutional layer's kernel stands for number of kernels on the concerned layer $M$, number of channels $C$, the width of a filter $W$, and the height of a filter $H$; the dimension of a fully-connected layer's weights means the number of preceding layer's neurons and the number of the concerned layers' neurons.}
\label{tab:archi}
\end{table}

In each experiment, we split the dataset into $3$ parts: $90\%$ of the training set is used for training the network, the remaining $10\%$ is used as validation set. 
During the training, we use both the training loss on training set and inference error-rate on the validation set as performance measurements. 
To evaluate the different trained models, we use the classification accuracy on the testing set as the evaluation protocol.

In order for all these benchmark to remain challenging, We didn't use any pre-processing, data-augmentation or unsupervised learning. 
We use binarized hard tangent~\cite{hubara2016bnn} function as activation function.
The ADAM adaptive learning rate method~\cite{kingma2014adam} is used while minimizing the square hinge loss with an exponentially decayed learning rate.
We also apply batch normalization to our networks, with a mini-batch of size $100$, $50$ and $50$ (separately for MNIST, CIFAR-10 and SVHN), to speed up the learning, and we scale the learning rate for each convolutional layer with a factor from Glorot's batch normalization~\cite{ioffe2015batchnorm}.
We train our networks for $300$ epochs on MNIST and CIFAR-10 datasets, and $200$ epochs on SVHN datasets.
The results are given in Sec.~\ref{sec:results}.

\subsection{Benchmark Result}
\label{sec:results}
\begin{table*}[ht]
\centering
\begin{tabular}{|l"c|c|c|}
\hline
Dataset                       & MNIST(CNN)  & CIFAR-10 & SVHN  \\ \thickhline
\multicolumn{4}{|c|}{No binarization (standard results)}       \\ \hline
Maxout Networks~\cite{goodfellow2013maxout} & 0.94\% & 11.68\%  & 2.47\%     \\ \hline
\multicolumn{4}{|c|}{Binarized Network}                        \\ \hline
BCNN(BinaryNet)~\cite{hubara2016bnn}		  & 0.47\% & 11.40\%  & 2.80\%     \\ \hline
\multicolumn{4}{|c|}{Binarized Network with Separable Filters} \\ \hline
BCNNw/SF Method 1 (this work)          & 0.48\% & 14.12\%  & 4.60\%    \\ \hline
BCNNw/SF Method 2 (this work)          & 0.56\% & 15.46\%  & 4.18\%   \\ \hline
\end{tabular}
\vspace{0.2cm}
\caption{Error Rate Comparison  on Different Datasets.
BCNNw/SF1 stands for our training method 1; BCNNw/SF2 denotes for our training method 2.}
\label{tab:results}
\end{table*}
\begin{figure}
	\centering
	\includegraphics[width=1.0\columnwidth]{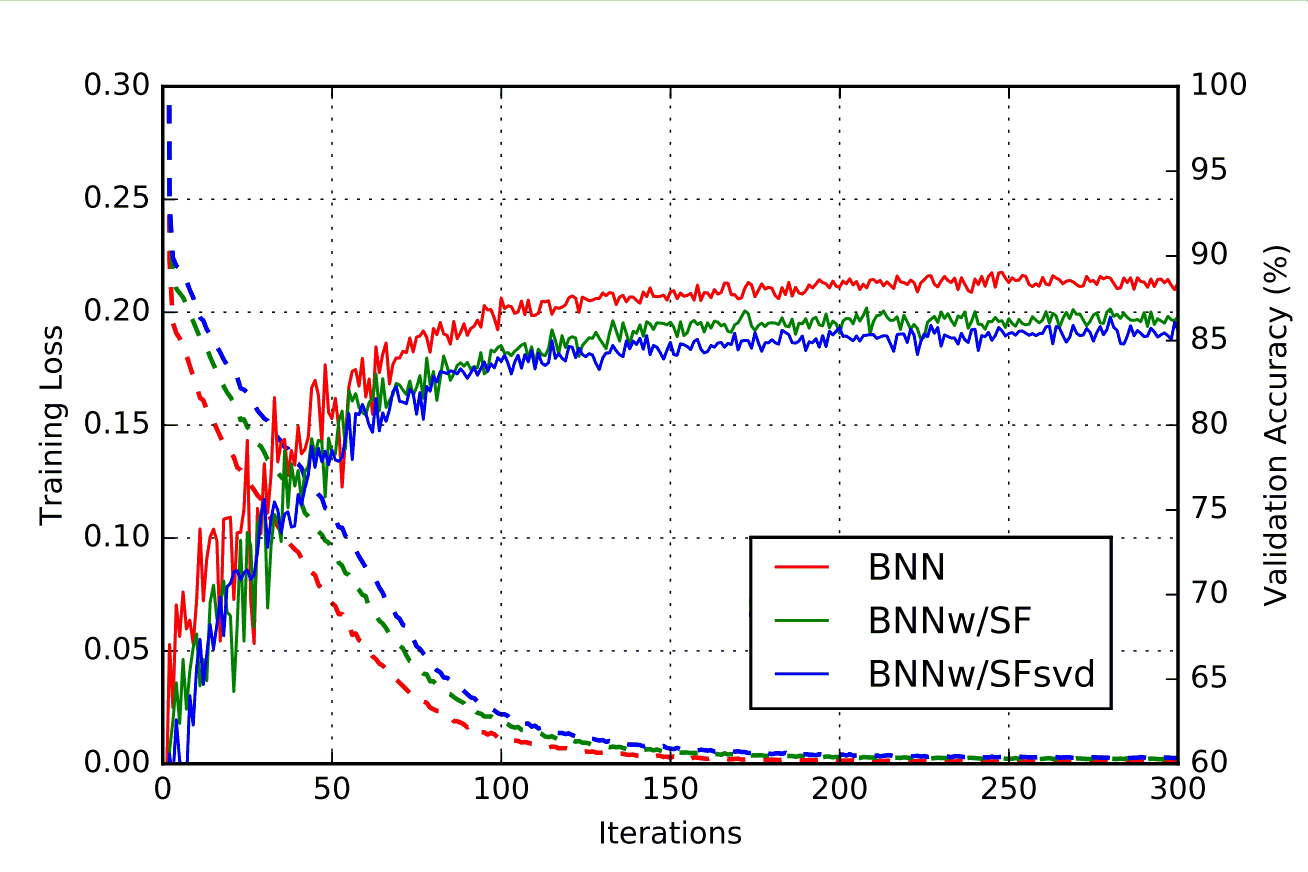}
	\caption{Learning Curves of ConvNets for BNN(red), BNNw/SF1(green) and BNNw/SF2(blue) on CIFAR-10 dataset. 
    The dotted lines represent the training costs(square hinge losses) and the continuous lines the corresponding validation accuracy.}
    \label{fig:learning_curve_cifar}
\end{figure}
Fig.~\ref{fig:learning_curve_cifar} depicts the learning curves on CIFAR-10 dataset.
There exists certain accuracy degradation if we compare BCNN with our methods due to a more aggressive noise.
By the end of the training phase, our method 1 yields an accuracy less than that of BCNN by roughly $2.72$\%, and the method 2 reaches a even more inferior accuracy.
For the sake of CIFAR-10's higher difficulty, the loss of accuracy meets our expectation.
We will discuss in detail the benefit of using exact gradient over the rank-1 approximation in next sub-section.

Tab.~\ref{tab:results} summarizes the experimental results in terms of error rate.
Compared with BNN~\cite{hubara2016bnn}, for the gray-scale manuscript number classification, both of our two training methods achieve a accuracy close to that of the binarized convolutional neural networks.
The difference is within 0.09\%.
It is noteworthy that our method 2 outperforms method 1 on SVHN by $0.42\%$ error rate.
For CIFAR-10 and SVHN, our methods are inferior to BCNN by a difference less than 2.72\% because we limit choices of filters from a number of $512$ to $32$, where the filter size is $3$x$3$. 
Since the performance degradation on CIFAR-10 is the largest, we implement a hardware accelerator in FPGA to inspect at what extent of hardware complexity can be improved with the sacrifice of the $2.72$\% accuracy loss. 
Sec.~\ref{sec:FPGA} provides the details and a comparison with a BCNN accelerator to demonstrate the benefits of BCNNw/SF.

\subsection{Scalability}
We also explore different sizes of networks to improve the accuracy and exam the scalability of BCNNw/SF.
Tab.~\ref{tab:archi_bigger} lists two additional larger models and an AlexNet-like model for CIFAR-10.
The wider one stands for a model with all numbers of kernels doubled, and the deeper one is a network including two extra convolutional layers.
Different from the models above, the AlexNet-like model includes three sizes of filters: $5$-by-$5$, $3$-by-$3$, and $1$-by-$1$.
Applying our rank-1 approximation on $5$-by-$5$ filter, we can get $64\%$ memory reduction.
\begin{table}[h]
\centering
\begin{tabular}{|c"c|c|c|}
\hline
Name       & Deeper      & Wider         & AlexNet-like\\ \thickhline
Input      & 3x32x32     & 3x32x32       & 3x32x32\\ \hline
Conv-1     & 128x3x3   & 256x3x3     & 96x5x5\\
Conv-2     & 128x3x3  & 256x3x3   & 256x5x5\\ \hline
Pooling    & \multicolumn{3}{c|}{2 x 2 Max Pooling} \\ \hline
Conv-3     & 256x3x3 & 512x3x3   & 512x3x3\\
Conv-4     & 256x3x3 & 512x3x3   & 512x3x3\\ \hline
Pooling    & \multicolumn{3}{c|}{2 x 2 Max Pooling} \\ \hline
Conv-5     & 512x3x3 & 1024x3x3  & 256x3x3\\
Conv-6     & 512x3x3 & 1024x3x3 & 512x1x1\\ \hline
Pooling    & \multicolumn{3}{c|}{2 x 2 Max Pooling}  \\ \hline
Conv-7     & 512x3x3 & -             & - \\
Conv-8     & 512x3x3 & -             & - \\ \hline
Pooling    & 2x2 Max Pooling & -       & - \\ \hline
FC-1       & 1024 & 1024     & 1024\\ \hline
FC-2       & 1024   & 1024     & 128 \\ \hline
FC-3       & 10     & 10       & 10   \\ \hline
\end{tabular}
\vspace{0.2cm}
\caption{ The 1st column shows a deeper model with two extra convolutional layers, and the 2nd column shows a widened network with all numbers of kernels doubled. The 3rd column is inspired by AlexNet to include 3 sizes of filters.}
\label{tab:archi_bigger}
\end{table}
\begin{figure}
	\centering
 	\includegraphics[width=1.0\columnwidth]{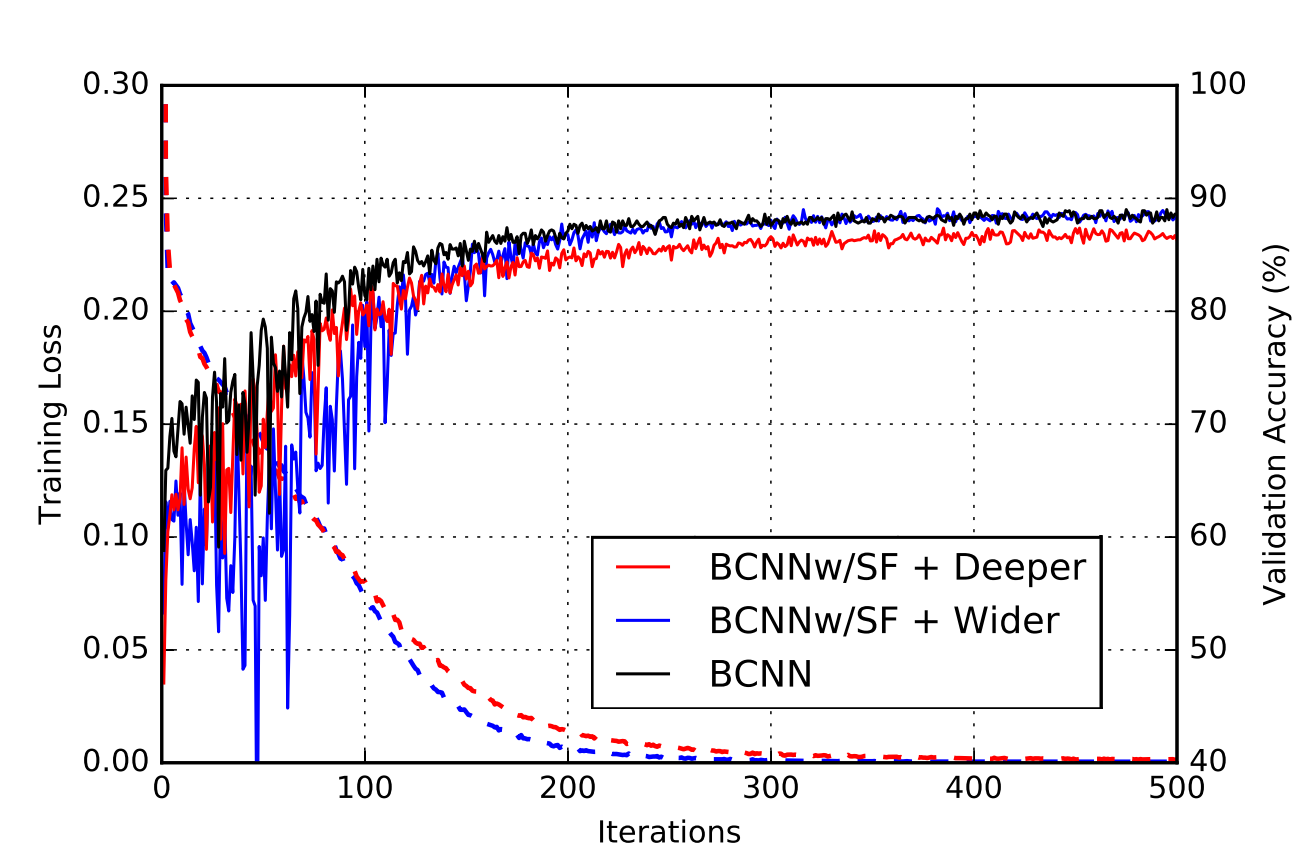}
	\caption{Learning Curves of ConvNets for BCNNw/SF with deeper Network(red), BCNNw/SF with wider Network(blue) and original BCNN on CIFAR-10 dataset.
    The dotted lines represent the training costs(square hinge losses) and the continuous lines the corresponding validation error rates.}
    \label{fig:learning_curve_all_500_cifar}
\end{figure}
We train the three bigger networks with our method 1, and Fig.~\ref{fig:learning_curve_all_500_cifar} shows the learning curves of the two enlarged models for CIFAR-10.
Since the number of trainable parameters has been increased, it requires more epochs to travel in the hypothesis space and reach a local minimum.
Therefore, we train these two bigger networks with $500$ epochs, and compare with BCNN(BinaryNet).
As shown in Fig.~\ref{fig:learning_curve_all_500_cifar} the wider one (blue) starts with largest ripple yet catch up the same performance as BCNN(black) does around the 175th epoch.

\begin{table}[t]
\centering
\begin{tabular}{|l"c|}
\hline
Dataset                       & CIFAR-10 \\ \thickhline
BCNN(BinaryNet)~\cite{hubara2016bnn}	  & 11.40\%  \\ \hline
\multicolumn{2}{|c|}{Binarized Network with Separable Filters (this work)} \\ \hline
BCNNw/SF Method 1          & 14.12\%  \\ \hline
BCNNw/SF Method 1 depper  & 14.11\%  \\ \hline
BCNNw/SF Method 1 wider   & \textbf{11.68\%}  \\ \hline
BCNNw/SF Method 1 AlexNet-like   & 15.1\%  \\ \hline
\end{tabular}
\vspace{0.2cm}
\caption{Classification Accuracy (Error Rate) of the three larger models.}
\label{tab:results_bigger}
\end{table}
Tab.~\ref{tab:results_bigger} lists the results on CIFAR-10 of the three bigger models as well as the CIFAR-10 results in Tab.~\ref{tab:results}.
The performance improvement of deeper network is very scarce since the feature maps experience the extra destructive max pooling layer as shown in Tab.~\ref{tab:archi_bigger}, which reduces the size of the first fully-connected layer, FC-1, and hence suppresses the improvement.
The wider network achieves $11.68\%$, which is very close to the performance of BCNN(BinaryNet).
The AlexNet-like model demonstrates that a model with $5$-by-$5$ filters sacrifices more accuracy to provide higher memory reduction.
In summary, the accuracy degradation of BCNNw/SF can be compensated by enlarging the size of network.

\subsection{Discussion}
In this section, we use the experimental results on CIFAR-10 as an example of detailed analysis.
We unpack the trained rank-1 filters and learning curves to gain a better understanding of the mechanism of BCNNw/SF.

\begin{figure}
	\centering
	\includegraphics[width=1.0\columnwidth]{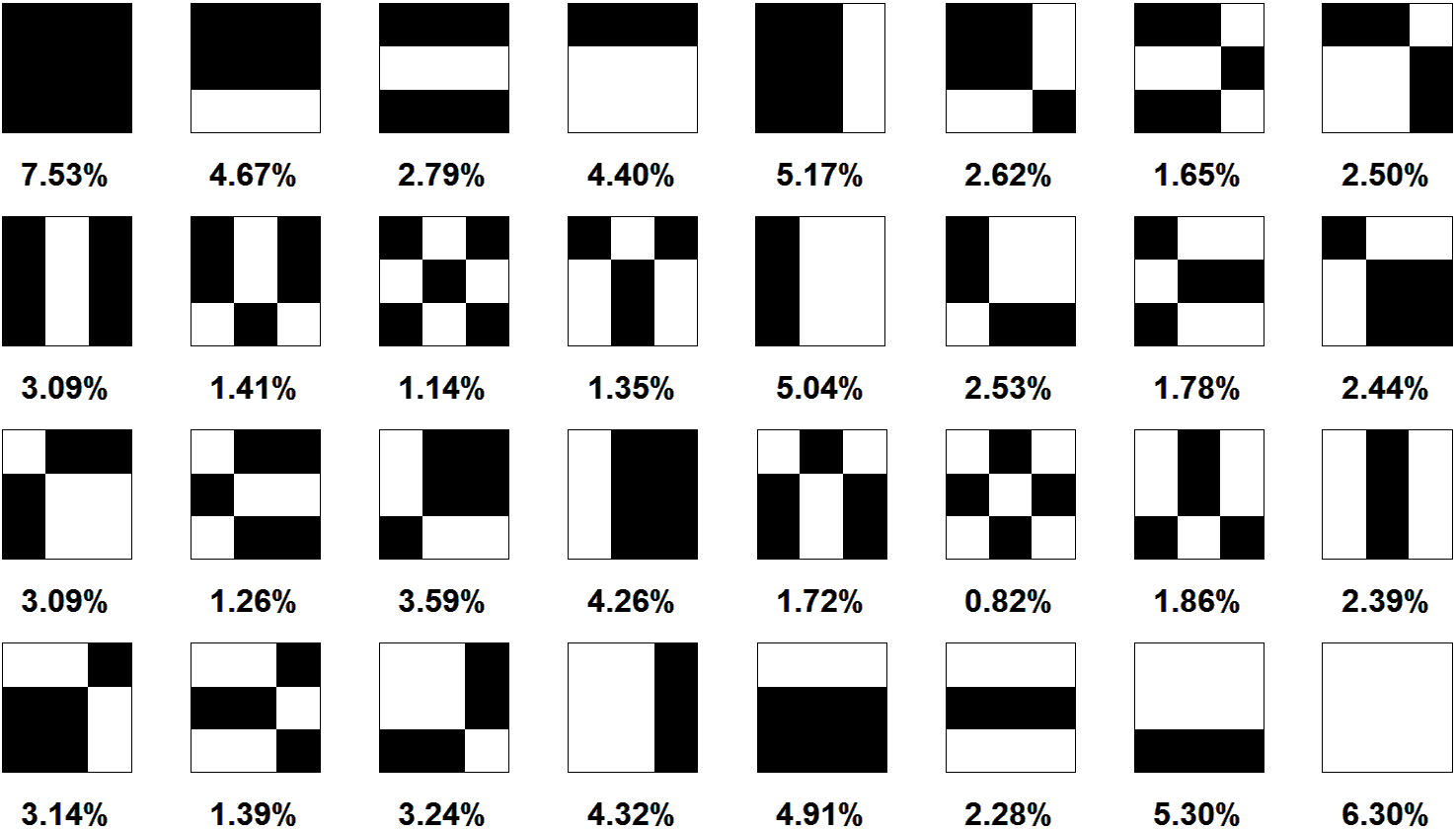}
    \vspace{0.05cm}
	\caption{Separable Filters \& Frequencies used in CIFAR Model}
    \label{fig:filter_draw}
\end{figure}
Fig.~\ref{fig:filter_draw} lists all the $32$ rank-1 filters and their frequency on CIFAR-10.
Although certain filters are rarely used, there is no filter forsaken.
In Fig.~\ref{fig:filter_draw} we can learn that the all-positive and all-negative filters are trained most frequently, and these two filters render the convolution to running-sum calculation with a sliding window.
As mentioned in Sec.~\ref{sec:design_choice},through the summation of separated convolution from a preceding layer, we can achieve the tangled linear combinations, which are essential to BCNNw/SF.

Unknowing the spectrum of the ripple, we apply Savitsky-Golay filter~\cite{savitzky1964sgfilter} to obtain the baseline of validation accuracy and, thereby, subtract the original accuracy with the baseline to get the ripple.
The window width of the Savitsky-Golay filter is $51$, and we use quadratic equation to fit the original learning curve.
All ripples are quantized into $100$ categories for the statistic analysis. 
\begin{table}
\centering
\begin{tabular}{|l"c|c|c|}
\hline
Statistics                       & mean & std & max \\ \thickhline
BCNN(BinaryNet)~\cite{hubara2016bnn} & 0.052 & 1.213 & 5.09  \\ \hline
BCNNw/SF Method 1          		 & 0.055 & 1.059 & 4.465 \\ \hline
BCNNw/SF Method 2  				 & 0.035 & 0.723 & 3.622 \\ \hline
\end{tabular}
\vspace{0.2cm}
\caption{The statistics of the ripples in terms of percent of error rate.}
\label{tab:ripple_stat}
\end{table}
Tab.~\ref{tab:ripple_stat} compares our method 1 and methods 2 with BCNN. 
All three statistic values of the method 2 are reduced.
The analytic gradient over the rank-1 approximation stabilizes the descending trajectory with more accurate gradient calculation.
Both BCNN and our method 1 rely on the gradient w.r.t. binarized filters to update all parameters due to the lack of analytic gradient w.r.t. real-valued filters.
However, it is also the rigorous gradient that limits the possibility to escape a local minimum on the error surface.
As we can see in Tab.~\ref{tab:results}, the results of our method 1 are closer to that of BCNN.
We use the trained binarized separable filter from our method 1 to implement a FPGA accelerator for CIFAR-10 in the following section.


\section{FPGA Accelerator}
\label{sec:FPGA}

\subsection{Platform and Implementation}
To quantify the benefits that BCNNw/SF can achieve for hardware BCNN accelerators, we created an FPGA accelerator for the six convolutional layers of the Courbariaux's CIFAR-10 network. Our accelerator is built from the open-source FPGA implementation in \cite{zhao-bnn-fpga2017}.
The dense layers were excluded as they are not affected by our technique. As BCNNw/SF is ideal for small, low-power platforms, we targeted a Zedboard with a Xilinx XC7Z020 FPGA and an embedded ARM processor.
This is a much smaller FPGA device compared to existing CNN FPGA accelerators~\cite{qiu2016cnn, suda2016opencl}.
We write our design in C++ and use Xilinx's SDSoC tool to generate Verilog through high-level synthesis.
We implement both BCNN and BCNNw/SF and examine the performance and resource usage of the accelerator with and without separable filters.

Our accelerator is designed to be small and resource-efficient; it classifies a single image at a time, and executes each layer sequentially.
The accelerator contains two primary compute complexes: \texttt{Conv1} computes the first (non-binary) convolutional layer, and \texttt{Conv2-5} is configurable to compute any of the binary convolutional layers.
Other elements include hardware to perform pooling and batch normalization, as well as on-chip RAMs to store the feature maps and weights.
Computation with the accelerator proceeds as follows.
Initially all input images and layer weights are stored in off-chip memory accessible from both CPU and FPGA.
The FPGA loads an image into local RAM, then for each layer it loads the layer's weights and performs computation.
Larger layers require several accelerator calls due to limited on-chip weight storage.  Intermediate feature maps are fully stored on-chip.
After completing the convolutional layers we write the feature maps back to main memory  and the CPU computes the dense layers.

We kept the BCNN and BCNNw/SF implementations as similar as possible, with the main difference being the convolution logic and storage of the weights.
For BCNN, each output pixel requires $3\times3=9$ MAC operations to compute.
For BCNNw/SF we can apply a $3$x$1$ vertical followed by a $1$x$3$ horizontal convolution, a total of $6$ MACs.
As the MACs are implemented by XORs and an adder tree, BCNNw/SF can potentially save resource.

In terms of storage, BCNN requires the $9$ bits to store each filter. Naively, BCNNw/SF requires $6$ bits, as each filter is represented as two $3$-bit vectors.
However, recall we only use rank-1 filters --- Eq.~\ref{eq:num_rank1_filters} shows that the number of unique $3\times3$ is $32$, meaning we can encode them losslessly with only $5$ bits.
A small decoder in the design is used to map the $5$-bit encodings into $6$-bit filters.

\subsection{Results and Discussion}
Table \ref{tab:fpga} compares the execution time and resource usage of the two FPGA implementations.
Resource numbers are reported post place and route, and runtime is wall clock measured on a real Zedboard.
We exclude the time taken to transfer the final feature maps from FPGA to main memory, as it is equal between the two networks; transfer time for the initial image and weights are included.
\begin{table}[htbp]
  \centering
    \begin{tabular}{lccc}
    \toprule
          & BCNN  & BCNNw/SF (this work) & $\delta$ \\
    \midrule
    \midrule
    Conv layer  & \multirow{2}{*}{0.949} & \multirow{2}{*}{0.652} & \multirow{2}{*}{-31.3\%} \\
    runtime (ms) &                        &                        & \\
    \midrule
    LUT        & 35255 & 36384 & +3.2\% \\
    FF         & 41418 & 41054 & -1.0\% \\
    Block RAM  & 94    & 78    &  -17.0\% \\
    DSP        & 8     & 8     &  0.0\% \\
    \bottomrule
    \end{tabular}%
    \vspace{0.1in}
  \caption{Comparison of performance and resource usage between BCNN and BCNNw/SF FPGA implementations. Runtime is for a single image, averaged over 10000 samples.}
  \label{tab:fpga}%
\end{table}%

Our experimental results show that BCNNw/SF achieves runtime reduction of 31\% over BCNN, which equates to a $1.46$X speedup.
This is due mostly to the reduction of memory transfer time of the compressed weight filters.
For similar reasons BCNNw/SF is able to save 17\% of the total block RAM (RAMs are used for both features and weights).
Look-up table (LUT) counts have increased slightly, due most likely to the additional logic needed to map the $5$-bit encodings to actual filters.
Overall, BCNNw/SF realizes significant improvements to performance and memory requirement with minimal logic overhead.

\section{Conclusion and Future Work}
\label{sec:concl}
In this paper, we proposed binarized convolutional neural network with Separable Filters (BCNNw/SF) to make BCNN more hardware-friendly.
Through binarized rank-1 approximation, 2D filters are separated into two vectors, which reduce memory footprint and the number of logic operations.
We have implemented two methods to train BCNNw/SF with Theano and verified our methods with various CNN architectures on a suite of realistic image datasets.
The first method relies on batch normalization to regularize noise, making it simpler and faster to train, while the second method uses gradient over SVD to make the learning curve more smooth and potentially achieves better accuracy.
We also implement an accelerator for the inference of a CIFAR-10 network on an FPGA platform.
With separable filters, the total memory footprint is reduced by $17.0\%$ and the performance of the convolution layers is improved by $1.46$X compared to baseline BCNN.

Integrating probabilistic methods~\cite{soudry2014ebp} to reduce the training time and exploring more elegant structures of networks~\cite{szegedy2016inception_v4} will be a promising direction for future works.

\vspace{-0.08in}
{
\scriptsize{
\bibliographystyle{abbrv}
\vspace{-0.05in}
\bibliography{BCNNwSF}  
}
}

\end{document}